# Don't Just Search, Understand: Semantic Path Planning Agent for Spherical Tensegrity Robots in Unknown Environments

Junwen Zhang[1], Changyue Liu[2], Pengqi Fu[1], Xiang Guo[3], Ye Shi[1], Xudong Liang[4],
Zhijian Wang[2], and Hanzhi Ma[1]

*Abstract*—Endowed with inherent dynamical properties that grant them remarkable ruggedness and adaptability, spherical tensegrity robots stand as prototypical examples of hybrid soft-rigid designs and excellent mobile platforms. However, path planning for these robots in unknown environments presents a significant challenge, requiring a delicate balance between efficient exploration and robust planning. Traditional path planners, which treat the environment as a geometric grid, often suffer from redundant searches and are prone to failure in complex scenarios due to their lack of semantic understanding. To overcome these limitations, we reframe path planning in unknown environments as a semantic reasoning task. We introduce a Semantic Agent for Tensegrity robots (SATPlanner) driven by a Large Language Model (LLM). SATPlanner leverages high-level environmental comprehension to generate efficient and reliable planning strategies. At the core of SATPlanner is an Adaptive Observation Window mechanism, inspired by the "fast" and "slow" thinking paradigms of LLMs. This mechanism dynamically adjusts the perceptual field of the agent: it narrows for rapid traversal of open spaces and expands to reason about complex obstacle configurations. This allows the agent to construct a semantic belief of the environment, enabling the search space to grow only linearly with the path length ($O(L)$) while maintaining path quality. We extensively evaluate SATPlanner in 1,000 simulation trials, where it achieves a 100% success rate, outperforming other real-time planning algorithms. Critically, SATPlanner reduces the search space by 37.2% compared to the A* algorithm while achieving comparable, near-optimal path lengths. Finally, the practical feasibility of SATPlanner is validated on a physical spherical tensegrity robot prototype.

## I. INTRODUCTION

Robustness and reliability are likely the first impressions most people have of robots: the rigid shells provide solid protection for the structure of robot, enabling it to maintain overall structural stability while performing complex tasks. However, as robotic applications expand, soft robots have demonstrated enormous potential in human-machine interaction, impact resistance, and environmental adaptability, attracting extensive attention from many scholars [1], [2], [3].

Among the various types of soft robots, the tensegrity robot is distinguished by its remarkable feats of dexterity and resilience [4]. Tensegrity robots are usually composed of rigid compressive elements (e.g., struts) and flexible tensile elements (e.g., cables), which are connected to create a compliant yet stable network. In this structure, the rigid compressive elements resemble the skeleton of the robot, while the flexible tensile elements act like tendons, with internal pre-tension to maintain structural integrity. This bio-inspired design, which combines rigid and flexible characteristics, endows tensegrity robots with numerous advantages such as flexibility, low weight, minimal material usage, high strength-to-weight ratios, and the capacity to absorb significant external loads. These features make them promising candidates for applications like space exploration and disaster relief [5], [6]. Among the various shapes and designs of tensegrity robots, the spherical tensegrity robot, illustrated in Fig. 1, stands out due to its highly symmetrical icosahedral geometry, making it an excellent mobile platform. Given the high degrees of freedom (DoF) in tensegrity robots, achieving efficient and stable motion is a critical area of research, known as locomotion [7], [8]. Furthermore, the challenge of enabling autonomous navigation for tensegrity robots based on a stable gait has been largely unexplored. Our work, therefore, focuses on the task of path planning for tensegrity robots in unknown environments.

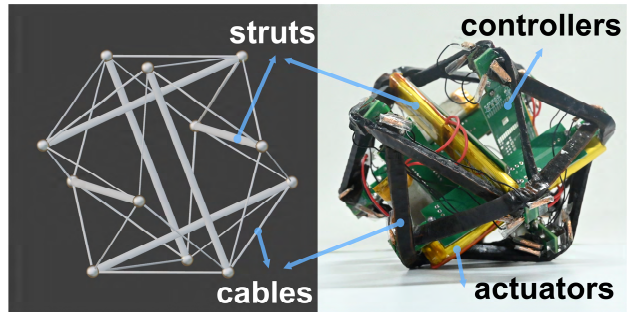

Fig. 1: *Left*: Spherical tensegrity robot in the blender simulation environment. *Right*: Physical spherical tensegrity robot prototype.

For any algorithm, path planning in unknown environments is fundamentally a trade-off between exploration and exploitation [9], [10]. Conventional algorithms typically rely on searching or sampling techniques to navigate the environment and locate the goal. A larger search space can imply a more thorough exploration, theoretically increasing the chances of finding an optimal path. However, this comes at the cost of higher resource consumption. Thus, an expansive search space is not always advantageous, as it can become computationally prohibitive in large-scale, complex environments. The key challenge is to strike a balance: achieving a short path length while exploring a minimal search space [11], [12]. Traditional methods, such as A* [13]and RRT* [14] commonly treat environmental perception in path planning as a purely digital grid or geometric problem. As a result, their optimality can be guaranteed only in known maps, while in practice they often suffer from significant redundancy in the search space. . Conversely, other methods like the Dynamic Window Approach (DWA) [15] can perform real-time path planning in unknown environments, but their reliance on local information makes them prone to failure or getting trapped in local minima. In summary, these methods lack a mechanism to understand the semantic information of the environment, guiding us to integrate semantic information into path planning task to enhance overall performance.

We propose SATPlanner driven by Large Language Model (LLM) as a solution. This agent leverages the powerful semantic analysis and reasoning capabilities of LLM while integrating functional modules to enable efficient and stable path planning in unknown environments. At the core decision-making level of path planning, we introduce an Adaptive Observation Window (AOW) mechanism designed based on the "fast" and "slow" thinking paradigms of LLMs. Through 1,000 simulation trials, we demonstrate that SATPlanner achieves a 100% success rate and reduces the search space by 37.2% compared to A*, while maintaining near-optimal path lengths. Finally, we conduct experiments on a physical tensegrity robot prototype to validate the practical feasibility of SATPlanner.

The main contributions of this paper can be summarized as follows:

1. We reframe path planning for spherical tensegrity robot in unknown environments from a geometric search problem into a semantic reasoning task driven by LLM.

2. We develop SATPlanner, a complete and robust end-to-end agent system for spherical tensegrity robot that performs efficient path planning by constructing and comprehending the semantic information of the environment.

3. We design an AOW mechanism that dynamically adjusts the granularity of perception based on environmental complexity, efficiently balancing the trade-off between exploration and planning.

4. We comprehensively validate the superior efficiency and sub-optimal performance of SATPlanner through extensive simulation experiments and a prototype demonstration.

## II. RELATED WORK

### A. Locomotion in Spherical Tensegrity Robots

Research on spherical tensegrity robots, actuated by up to 24 cables, has mainly focused on developing control strategies for rolling locomotion. Data-driven methods, including evolutionary algorithms [16] and Bayesian optimizers [17], have been used to generate control strategies. Supervised learning has also been used to train a contextual policy for directed rolling on flat terrain [18]. Reinforcement learning is also regarded as a promising method and has been extensively studied [19], [20] with Guided Policy Search (GPS) [21] successfully generating efficient rolling gaits. These advances have extended applications from basic locomotion to tasks such as slope climbing [22], [23]. However, higher-level tasks beyond locomotion like path planning remain underexplored, with only limited efforts such as a modified RRT* for trajectory planning [24].

### B. Classical and Real-Time Path Planning Algorithms

Path planning is a crucial task in artificial intelligence and robotics, with numerous algorithms developed to address its challenges. Searching- and sampling-based methods typically operate on known maps. These methods employ different search or sample strategies to construct state estimations of the environment, thereby accomplishing path planning. Widely used algorithms in this category include Dijkstra algorithm, A* algorithm [13] and RRT* algorithm [14].

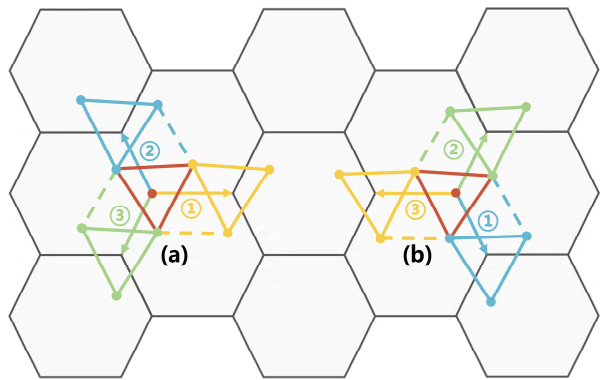

Fig. 2: The two rolling gaits of spherical tensegrity robots. The combination of (a) and (b) gaits enables the motion trajectory of robot to approximate a regular hexagonal grid.

Unlike most searching- and sampling-based methods that require a priori maps, real-time path planning algorithms, like the Dynamic Window Approach (DWA) [15] and Artificial Potential Field (APF) [25], are designed to operate in unknown environments. Recently, post-training models have emerged as an effective approach for path planning, with studies employing DRL framework with context-aware mechanisms to achieve mapless navigation [26].

### C. Large Language Models for Robotic Path Planning

LLMs have recently demonstrated powerful capabilities in multiple fields. Their enhanced reasoning capabilities now allow them to perform functions analogous to human thought and decision-making processes, making them suitable for complex tasks. However, the application of LLMs in path planning for robotics is still in its early stage [27]. For instance, some studies utilize LLMs as optimizers to generate waypoints, which are then connected by the A* algorithm to form a complete path [28]. Others employ LLMs as translators to convert natural language into Signal Temporal Logic (STL), subsequently combined with STL-based methods for motion planning [29]. Additionally, some approaches directly construct path planning agents using a Video Language Model for autonomous navigation [30].

## III. ROLLING GAIT OF SPHERICAL TENSEGRITY ROBOT

As shown in Fig. 1, the spherical tensegrity robot consists of 20 triangles forming a stable icosahedral structure, including 12 isosceles triangles and 8 equilateral triangles. Analysis of its rolling mechanism indicates that motion can be decomposed into a two-step continuous process. The initial state is defined with an equilateral triangle as the base. Rolling begins by contracting one edge of this base triangle, which shifts the center of gravity of the robot and causes it to roll onto an adjacent isosceles triangle. Subsequently, by contracting the edge opposite to the isosceles triangle, the robot rolls onto another equilateral triangle adjacent to the current one. This two-step rolling gait enables the robot to transition between equilateral triangles, as illustrated in Fig. 2 with two opposite transition patterns.

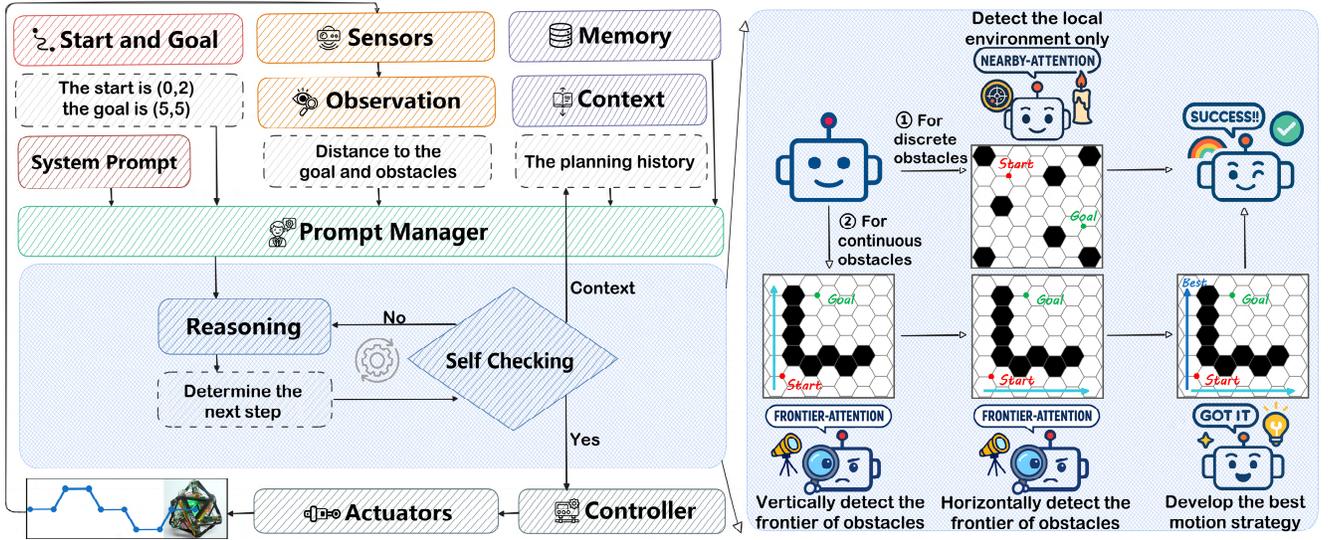

Fig. 3: SATPlanner Architecture and Adaptive Observation Window mechanism. *Left*: the architecture of SATPlanner is illustrated, showing the various modules that constitute the agent system and their collaborative workflow. *Right*: The right side presents a schematic of the AOW mechanism, demonstrating how SATPlanner applies different strategies in two distinct types of environmental scenarios.

Follow this pattern, the spherical tensegrity robot exhibits rolling locomotion that approximates a hexagonal mode, which is difficult to simulate accurately using simple square grids or curves. Therefore, hexagonal grids are employed to build the simulation environment in this work to ensure that the simulated rolling trajectories are more readily transferable to real-world applications. The detailed design is provided in Section V. Building on this rolling gait, a semantic path planning agent for spherical tensegrity robots in unknown environments is further developed.

## IV. PROPOSED PATH PLANNING METHOD

### A. Formulation of Path Planning in Unknown Environments

We formulate the task of path planning for spherical tensegrity robot in unknown environments as a Partially Observable Markov Decision Process (POMDP). A POMDP can be defined by a 7-tuple $M = (S, A, T, R, \Omega, O, \gamma)$, where:

**State Space** $S$: a state $s \in S$ includes the exact position of the robot, the complete layout of all obstacles, and the precise coordinates of the goal. This global state is unobservable to the agent.

**Action Space** $A$: derived from the gait of the spherical tensegrity robot, the action space initially consists of three discrete motion directions, each corresponding to one of the equilateral triangles adjacent to the base equilateral triangle. In our agent architecture, however, the position of the robot is represented on a hexagonal grid for clearer task definition. The controller then maps grid-based rolling paths to actual physical trajectories. Accordingly, the action space of the agent is defined as moving to one of the six adjacent cells in the hexagonal grid.

**Transition Function** $T$: $T(s'|s,a)$ defines the probability of transitioning to a new state $s'$ after executing action $a$ in state $s$. This function is governed by the underlying kinematics of the robot and the physical dynamics of the environment. In this task, performing an action $a \in A$ changes the position of the robot, thereby updating the state space.

**Reward Function** $R$: $R(s,a)$ represents the immediate reward received when the agent takes action $a$ in state $s$. In our framework, the reward function is implicit and defined by three criteria: 1) making progress toward the goal, 2) successfully navigating around consecutive obstacles, and 3) performing correct backtracking. The **discount factor** $\gamma$ balances immediate and future rewards. For the agent, this factor is also implicit and may adapt dynamically across different stages of the path planning process.

**Observation Space** $\Omega$: an observation $o \in \Omega$ is represented as a natural language prompt with the structure:

$$o = o_{system} + o_{memory} + o_{sensor} \quad (1)$$

Here, $o_{system}$ represents the system prompt, $o_{memory}$ encodes the historical memory of the agent during path planning, and $o_{sensor}$ represents LiDAR-based perception of the surrounding environment.

**Observation Function** $O$: $O(o_t | s', a_{t-1})$ defines the probability that the agent receives a specific language observation $o_t$ after the environment transitions to a new state $s'$. In our task, this function abstracts the entire process from perceiving the physical world through sensors, to the Prompt Manager integrating historical information, and ultimately generating a complete text prompt.

Within this POMDP framework, SATPlanner aims to learn an optimal policy $\pi^*(a_t | b_t)$ that maps the current belief state $b_t$ to an optimal action $a_t$. Traditional methods, such as reinforcement learning, typically approximate $\pi^*$ by explicitly learning a value function or a policy network. Based on the powerful reasoning and semantic understanding capabilities, the LLM, prompted with the current observation and memory, acts as a powerful heuristic policy function. It leverages its pre-trained semantic understanding to construct a semantic belief of the current state and perform zero-shot cost estimation, effectively approximating a solution to this POMDP without explicit value function calculation or policy training

*B. SATPlanner Architecture*

The architecture of SATPlanner is illustrated in Fig. 3. The system begins with the **System Prompt**, which initializes the SATPlanner by defining the **Start and Goal** and path planning task. This step configures the LLM from a general assistant to a domain-specific agent tailored to path planning in unknown environments.

Environmental perception is provided by the **Sensors** module. LiDAR is employed to perceive the surroundings. The raw LiDAR data are processed by the system and converted into a formalized "**Observation**" text before fed into the LLM.

Path planning is a long-horizon, sequential process, where equipping the LLM with global planning capabilities is crucial for success. However, the limited context window of LLMs can lead to model hallucinations when the dialogue context becomes too long. In path planning, such deviations can cause loops or broken paths, degrading performance. Therefore, we introduce the **Memory** module to enhance long- and short-term memory from two angles: intrinsic (model-side) and external (data-side).

On the model side, a **Context** management and summarization mechanism is employed. After a fixed number of dialogue turns, the model triggers a self-summarization routine that produces a state summary from the current context, including 1) current position, 2) destination, 3) global obstacle list, 4) path history, 5) verified obstacle prediction patterns, 6) hypotheses for obstacle prediction in unexplored areas, and 7) key decision points. This summary consolidates key historical information from the planning process and captures salient semantic insights for self-reflection, thereby equipping the LLM with effective, self-maintained long-term memory.

For external memory, a local memory database indexed by grid coordinates is introduced, storing only basic environmental information around each location. As the agent moves, the database is continuously updated, and at each step, relevant memories exceeding a predefined threshold are retrieved based on the current position to support local planning.

These two mechanisms, addressing short-term memory and long-term memory, respectively, enhance the capabilities of LLM in both local and global planning, leading to more stable and efficient task completion.

The outputs from the System Prompt, Sensors, and Memory modules are integrated in the **Prompt Manager**, which compiles them to generate the final structured natural-language inputs for the LLM.

During the **Reasoning** process, the LLM will carefully digests the input prompt to make a comprehensive decision on the next step. However, it may produce logical inconsistencies, a phenomenon known as "hallucination". To mitigate this, a **Self-Check** module is introduced. Here, we primarily conduct semantic verification of the chain of thought of the LLM from two perspectives: first, determining path topology connections to prevent broken paths; second, identifying obstacle locations to avoid misjudgments caused by hallucinations. This module enables the model to validate its own output before generation, thereby reducing logical errors and enhancing the operational robustness of SATPlanner.

Finally, the **Controller** bridges the decision-making LLM and the physical actuation of the robot. First, it converts the output of LLM into a physical rolling trajectory. This trajectory is then matched to the appropriate control logic based on the specific gait patterns of the spherical tensegrity robot. Second, the module handles robot rolling control by mapping the control logic to the corresponding cables and sending control signals to the relevant **Actuators** via Local Area Network (LAN). This process drives cable deformation, enabling the robot to roll along the predetermined trajectory.

In summary, building on the powerful semantic reasoning and decision-making capabilities of the LLM, SATPlanner integrates several key modules. The Sensors module provides the LLM with environmental perception, the Memory module enhances its local and global planning abilities, and the Self-Check module improves the robustness of its reasoning. The Controller module allows the LLM to focus on high-level path planning by abstracting away low-level control. Collectively, these modules enable the agent to complete the path planning task with greater efficiency and stability.

*C. Adaptive Observation Window Mechanism*

LLMs have demonstrated remarkable versatility and broad applicability for general tasks. However, their performance can be sub-optimal when applied to specialized, domain-specific tasks. In the context of path planning in unknown environments, relying solely on the native reasoning and decision-making capabilities of LLM can easily lead the agent to local minima, as decisions must be made based on limited sensory information in unknown environments. To address this, we propose the AOW mechanism. The AOW mechanism formalizes the sensory information received by the LLM into an observation window and performs path planning by reasoning based on the information in the window. Importantly, the mechanism enables the SATPlanner to adaptively adjust the window size according to environmental complexity.

The mechanism can be illustrated through two common path planning scenarios:

**1) Discrete obstacles**. In relatively simple environments with discrete obstacles, the observation window size is set to 1. This allows the LLM to process only the environmental information within the six hexagonal cells adjacent to its current position, thereby enabling a rapid decision-making. As shown in the right of Fig. 4, the agent adopts a nearby-attention approach, detecting only the local environment to make decisions efficiently.

**2) Continuous obstacles**. In complex environments with continuous obstacles, the LLM expands its observation window to obtain a more comprehensive understanding of the surroundings. As depicted in the right of Fig. 4, during this expansion, the LLM continuously explores along the obstacle distribution until both the positive and negative boundaries are determined. This exploration unfolds over multiple dialogue turns, simulating the perception and slow-thinking process of LLM. Once the boundaries are obtained, the LLM leverages the environmental information and the destination to select an optimal direction and generate the next motion strategy, allowing the robot to navigate the complex region successfully. This frontier-attention approach enables the agent to develop a global awareness of obstacles to improve path planning.

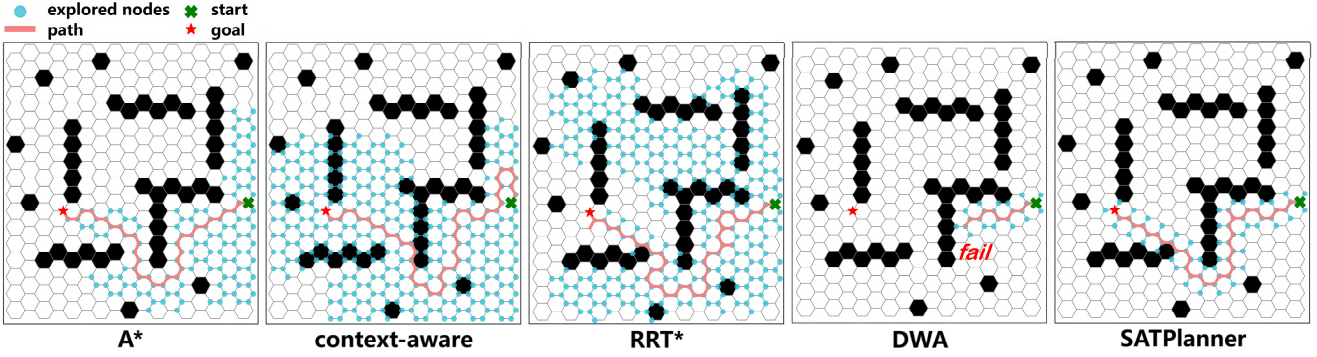

Fig. 4: Experimental simulation environment and comparison of SATPlanner with other methods under Path Length and Search Space metrics, including searching-based methods (A*), sampling-based methods (RRT*), learning-based methods (context-aware), and real-time path planning algorithms (DWA).

The AOW mechanism shifts the focus of the SATPlanner into two new tasks: first, dynamically customizing the window size based on current environmental information, and second, formulating motion strategies for continuous obstacles. This shift in decision point allows LLM to concentrate on its primary strength—establishing semantic beliefs about the environment. The integration of semantic information, in turn, enhances the precision of local planning, leading to a significant improvement in overall planning performance.

*D. Search Space Complexity Analysis for SATPlanner*

The primary bottleneck of the traditional A* algorithm lies in its inability to integrate a high-level, semantic understanding of the environment into its search strategy, which results in significant redundancy in search space. In contrast, our AOW mechanism in SATPlanner leverages the semantic consciousness of LLMs to perform aggressive pruning of the search space. In this section, we construct two idealized environmental scenarios to analyze the search space sizes of A* and SATPlanner, further demonstrating the effectiveness of the AOW mechanism in search space reduction.

The A* algorithm uses a cost function each time it expands a node.

$$f(n) = g(n) + h(n) \quad (2)$$

where $g(n)$ represents the exact cost from the start $s$ to node $n$ and $h(n)$ describes the Euclidean distance between node $n$ and the goal $g$.

Let $C*$ denote the optimal path cost. Due to the optimality of the A* algorithm, the algorithm inevitably expands all nodes satisfying $g(n) + h(n) < C*$ and may also expand several nodes for which $g(n) + h(n) = C*$.

**Simple environment.** Considering an environment with discrete and sparse obstacles. Based on the AOW mechanism, SATPlanner fixes the observation window size to 1, meaning it only expands adjacent grid nodes. Since observation windows overlap in consecutive steps, for a planned path of total length $L$, there exists a constant $c_1 \in [2,6]$ such that

$$S_{LLM} \leq c_1 L + O(1) = \Theta(L) \quad (3)$$

For the A* algorithm, in the ideal case, the heuristic causes it to preferentially expand nodes with a lower cost. Therefore, there exists a constant $c_2$ such that

$$S_{A^*} = c_2 L = \Theta(L) \quad (4)$$

In the worst case, however, obstructions from the sparse obstacles force A* to expand a larger number of nodes, and its search area approximates a circular region with a diameter equal to the path length $L$. In this case, the search space of A* is approximately

$$S_{A^*} = \Theta(L^2) \quad (5)$$

Thus, in simple environments, the search space of SATPlanner consistently scales as $\Theta(L)$, whereas the search space of the A* method ranges between $\Theta(L)$ and $\Theta(L^2)$.

**Complex environment.** We now consider an environment with continuous obstacles. The AOW mechanism drives the SATPlanner to adaptively increase the size of the observation window. Let the width of the continuous obstacle be $w$ and its length $h$. SATPlanner explores the frontier of obstacles in positive and negative directions until both boundaries of obstacles are detected. Therefore, the size of its search space is dependent on the perception length in two directions. Therefore, there exists a constant $c_3 \in [2,6]$ such that

$$S_{\text{LLM}} = c_3(h+w) = \Theta(h+w) \quad (6)$$

TABLE I. COMPARISON OF RESULTS FROM 1.000 INDEPENDENT EXPERIMENTS FOR ALL ALGORITHMS

| Method | | Success Rate ↑ | Path Length ↓ | Search Space ↓ | Weighted Search Space ↓ |
|---|---|---|---|---|---|
| Search-Based algorithms | A* | **100%** | **25.97** | 101.33 | 2631.19 |
| | Dijkstra | **100%** | **25.97** | 251.04 | 6518.76 |
| Sample-Based algorithms | RRT* | **100%** | 36.52 | 142.42 | 5201.39 |
| | BIT* | 81.9% | 28.33 | 412.31 | 11681.1 |
| Real-time path planning algorithms | DWA | 54.4% | 26.31 | / | / |
| | APF | 42.8% | 27.02 | / | / |
| Learning-Based algorithms | Context-aware | 83.6% | 63.67 | 344.83 | 21955.33 |
| Proposed Method | **SATPlanner** | **100%** | 26.88 | **63.59** | **1709.6** |

TABLE II. COMPARISON RESULTS OF ABLATION EXPERIMENT

| Method | Success Rate | Path Length | Search Space |
|---|---|---|---|
| SATPlanner | **100%** | **34.3** | **78.39** |
| no memory | 83.3% | 35 | 81.36 |
| no AOW | 94.4% | 40.47 | 70.66 |
| no self-check | 93.6% | 34.8 | 79.33 |

For the A* algorithm, however, due to its lack of semantic decision-making capability, continuous obstacles will block all nodes within the enclosed space, causing them to satisfy $g(n)+h(n) \leq C*$, so A* will expand all nodes within this region. Therefore, there exists a constant $c_4$ such that

$$S_{A^*} = c_4 * h * w = \Theta(h*w) \quad (7)$$

Therefore, when $w$ and $h$ increase by the same order, we can conclude that

$$S_{\text{LLM}} < S_{A^*} \quad (8)$$

Based on the two scenarios above, it is evident that SATPlanner leverages the AOW mechanism to adaptively adjust the observation window size for semantic pruning, which effectively reduces the search to a linear space, whereas the A* algorithm requires an extensive two-dimensional search to guarantee path optimality. Therefore, under identical conditions, the search space of SATPlanner is consistently smaller than that of A* algorithm and this advantage becomes increasingly pronounced as the map size grows.

## V. EXPERIMENT

### A. Experimental Setup

**Datasets.** To accommodate the specific gait characteristics of the spherical tensegrity robot, we develop an experimental environment using a hexagonal grid map to validate our path planning algorithms. As shown in Fig. 4, the map is composed of regular hexagonal cells; obstacles are represented as black hexagonal cells, and the robot moves along the vertices of the hexagonal grid. We assemble a dataset of 100 randomly generated maps. For each map, 10 random start–goal pairs are sampled. Consequently, each algorithm is evaluated over a total of 1,000 path planning trials.

**Large Language Model.** In SATPlanner, we employ the latest Large Reasoning Model from OpenAI and Deepseek (OpenAI-o1 [31] and Deepseek-R1 [32]) for their balance of robustness and cost-effectiveness.

**Evaluation Metrics.** To evaluate performance comprehensively, we adopt four quantitative metrics: 1) **Success Rate**: the ratio of trials in which a valid path is successfully found. 2) **Path Length**: the average length of the paths in successful trials. 3) **Search Space**: The size of the space explored by the algorithm during planning. 4) **Weighted Search Space**: The product of the Search Space and Path Length. This combined metric provides a more holistic assessment of the overall balance between efficiency (search space) and optimality (path length).

**Evaluation Baseline.** For benchmarking, SATPlanner is compared against seven representative algorithms drawn from four categories: search-based methods (A* and Dijkstra), sampling-based methods (RRT* and BIT* [33]), learning-based methods (a context-aware reinforcement learning model [26]), and real-time path planning methods (Dynamic Window Approach (DWA) [24] and Artificial Potential Field (APF) [25]).

### B. Experimental Results

TABLE I presents the results of the experimental comparison. As a planner in unknown environments, SATPlanner achieves a 100% success rate across 1,000 independent trials, demonstrating a substantial advantage over other real-time path planning algorithms (54.4% and 42.8%). Regarding Search Space metrics, SATPlanner achieves the best performance across all baselines. It also achieves a significant 37.2% reduction in search space compared to the A* algorithm, even achieving improvements of up to several times compared to other methods in the baseline. This result is consistent with the conclusion presented in Section IV. As for the Weighted Search Space metrics, SATPlanner again ranks best among all baselines. It achieves a 35% improvement over A* and up to 12.8X improvement over learning-based method in the joint metric combining path length and search space.

Fig. 4 presents experimental results of SATPlanner alongside four other path planning algorithms in the same map environment. Except for the real-time path planning algorithm DWA, all other algorithms successfully complete the path planning. Regarding path length, SATPlanner achieves a sub-optimal path close to A* algorithm. Furthermore, the search space of SATPlanner is significantly smaller than that of A*, with this advantage even more pronounced compared to RRT* and learning-based methods. This further demonstrates the outstanding overall performance of SATPlanner.

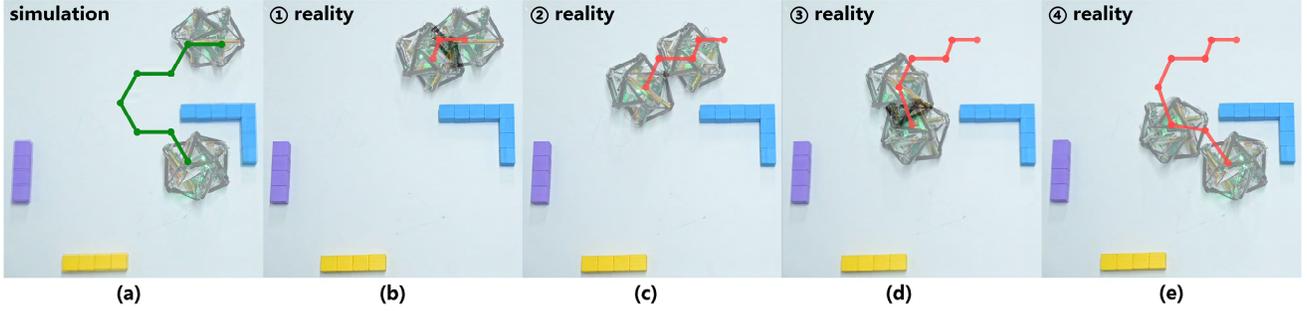

Fig. 5: (a) shows the path of robot in the simulated environment. (b)-(e) describe the sequential process of path planning performed by the robot.

These comparison results highlight the significant contribution of SATPlanner to advancing the study of the trade-off between searching and planning in path planning task. Specifically, SATPlanner demonstrates that near-optimal paths can be obtained while maintaining a substantially smaller search space.

*D. Ablation Studies*

The superior overall performance of SATPlanner stems from the synergistic combination of its functional modules with the powerful reasoning capabilities of the LLM. In this section, we conduct ablation experiments to investigate the contribution of individual components. Specifically, we evaluate the performance of SATPlanner after removing the Memory module (no memory), the Adaptive Observation Window mechanism (no AOW) and Self-Check module (no self-check), respectively.

As shown in TABLE II. The experimental results indicate that removing the Memory module impairs the global planning performance of SATPlanner, ultimately causing the success rate to drop to 83.3%. Similarly, without the AOW mechanism, the observation window of agent defaults to a fixed size of 1, resulting in a smaller search space compared to the full system. However, this diminished local planning ability also causes the agent to frequently enter local loops, which severely degrades path quality and success rate. Furthermore, removing the self-check module weakens the robustness of SATPlanner, leading to broken paths in some experiments, and the success rate declines to 93.6%.

*E. Hardware Implementation in a Real-World Scenario*

A spherical tensegrity robot with an untethered control system, illustrated in Fig. 1, is constructed to validate the effectiveness of SATPlanner. Path planning experiments is conducted on a flat laboratory surface, with the environment depicted in Fig. 5. The robot is initialized at a designated start point, and the experiment aims to have it navigate around obstacles toward the goal under SATPlanner guidance. Fig. 5(a) illustrates the robot path generated by SATPlanner in the simulated environment, while in Fig. 5(b)-(e), the actual trajectory of the robot has been recorded by connecting its geometric center over time. The results show that the robot successfully avoids obstacles and reaches the destination, with its actual trajectory closely aligning with the simulated path, demonstrating the applicability and excellent sim-to-real transferability of proposed SATPlanner for path planning of real tensegrity robots in practical scenarios.

## VI. CONCLUSION AND FUTURE WORK

In this work, we innovatively provide an efficient solution for path planning of spherical tensegrity robots in unknown environments. We present SATPlanner, a novel end-to-end LLM-driven agent that successfully and efficiently performs path planning in unknown environments. By reframing the problem from a geometric search to a semantic reasoning task, SATPlanner overcomes key limitations of traditional planners. SATPlanner leverages the powerful semantic understanding and reasoning capabilities of LLMs. Through the coordinated integration of multiple modules, SATPlanner gains robust and effective global planning capabilities. Furthermore, an Adaptive Observation Window mechanism is introduced to specifically enhance the local planning capabilities of SATPlanner and enable a linear relationship in complexity between the search space and the path length ($O(L)$). This holistic design enables the SATPlanner to complete path planning tasks with high quality. SATPlanner achieves near-optimal paths with a 100% success rate and a 37.2% reduction in search space compared to the optimal A* algorithm in extensive simulations, with performance validated on a physical robot. Our findings demonstrate that leveraging the semantic understanding capabilities of LLMs allows for the generation of efficient, sub-optimal paths from a remarkably small search space, marking a significant step towards more intelligent and adaptable autonomous navigation.

As for limitations, the primary limitation of SATPlanner lies in its real-time performance. At each step, the robot must await the output of the LLM before executing a movement. This process introduces significant latency due to the model inference time, causing our method to lag behind other algorithms in real-time responsiveness. Future work will focus on mitigating this latency by utilizing smaller, more responsive LLMs. The goal is to enhance the real-time capabilities of the agent and achieve a more comprehensive overall performance.